\documentclass[sigconf]{acmart}

\usepackage{subcaption}

\usepackage[utf8]{inputenc}
\usepackage[T1]{fontenc}
\usepackage{hyphenat}
\usepackage{xspace}
\usepackage{amsmath}
\usepackage{amsfonts}
\usepackage{hyperref}
\usepackage{url}
\usepackage{booktabs}
\usepackage{multirow}
\usepackage{makecell}
\usepackage{caption}
\usepackage{minibox}
\usepackage{bbm}
\usepackage{graphicx}
\usepackage{balance}
\usepackage{mathtools}
\usepackage{color}
\usepackage{marvosym}
\usepackage{ifthen}
\usepackage{textcomp}
\usepackage{enumitem}
\usepackage{verbatim}
\usepackage{algorithm}
\usepackage{numprint}
\usepackage{balance}

\usepackage{amsthm}
\theoremstyle{plain}

\newcommand{\chatoDisplayMode}[1]{#1}

\definecolor{MyRed}{rgb}{0.6,0.0,0.0} 
\definecolor{MyBlack}{rgb}{0.1,0.1,0.1} 
\newcommand{\inred}[1]{{\color{MyRed}\sf\textbf{\textsc{#1}}}}
\newcommand{\frameit}[2]{
  \begin{center}
  {\color{MyRed}
  \framebox[.9\columnwidth][l]{
    \begin{minipage}{.85\columnwidth}
    \inred{#1}: {\sf\color{MyBlack}#2}
    \end{minipage}
  }\\
  }
  \end{center}
}

\newcommand{\note}[2][]{\chatoDisplayMode{\def\@tmpsig{#1}\frameit{{\Pointinghand} Note}{#2\ifx \@tmpsig \@empty \else \mbox{ --\em #1}\fi}}}
\newcommand{\todo}[2][]{\chatoDisplayMode{\def\@tmpsig{#1}\frameit{{\Writinghand} To-do}{#2\ifx \@tmpsig \@empty \else \mbox{ --\em #1}\fi}}}

\newcommand{\abbrevStyle}[1]{#1}

\newcommand{\cf}{\abbrevStyle{cf.}\xspace}

\newcommand{\Secref}[1]{Sec.~\ref{#1}}

\newcommand{\Tabref}[1]{Table~\ref{#1}}
\newcommand{\Figref}[1]{Fig.~\ref{#1}}

\newcommand{\xhdr}[1]{\vspace{1.7mm}\noindent{{\bf #1.}}}

\newcommand{\textcite}[1]{\citeauthor{#1} \shortcite{#1}}

\newcommand{\hide}[1]{}

\makeatletter
\newcommand{\iffont}[2]{\ifthenelse{\equal{\f@family}{#1}}{#2}{}}
\makeatother

\AtBeginDocument{%
  }

\copyrightyear{2024}
\acmYear{2024}
\setcopyright{acmlicensed}\acmConference[WSDM '24]{Proceedings of the 17th ACM International Conference on Web Search and Data Mining}{March 4--8, 2024}{Merida, Mexico}
\acmBooktitle{Proceedings of the 17th ACM International Conference on Web Search and Data Mining (WSDM '24), March 4--8, 2024, Merida, Mexico}
\acmPrice{15.00}
\acmDOI{10.1145/3616855.3635825}
\acmISBN{979-8-4007-0371-3/24/03}

\newcommand{\modelname}{meta-model}
\newcommand{\framework}{FORC}

\begin{document}



%
\title{Fly-Swat or Cannon? Cost-Effective Language Model Choice via Meta-Modeling}

\author{Marija \v{S}akota}
 \affiliation{%
   \institution{EPFL}
   \country{Switzerland}}
\email{marija.sakota@epfl.ch}

\author{Maxime Peyrard}
 \affiliation{%
   \institution{EPFL}
   \country{Switzerland}}
\email{maxime.peyrard@epfl.ch}

\author{Robert West}
 \affiliation{%
   \institution{EPFL}
   \country{Switzerland}}
\email{robert.west@epfl.ch}

\newcommand\marija[1]{\textcolor{teal}{[Marija: #1]}}

\renewcommand{\shortauthors}{\v{S}akota, Peyrard, and West}

\begin{abstract}
Generative language models (LMs) have become omnipresent across data science.
For a wide variety of tasks, inputs can be phrased as natural language prompts for an LM, from whose output the solution can then be extracted.
LM performance has consistently been increasing with model size---but so has the monetary cost of querying the ever larger models.
Importantly, however, not all inputs are equally hard: some require larger LMs for obtaining a satisfactory solution, whereas for others smaller LMs suffice.
Based on this fact, we design a framework for cost-effective language model choice, called \textit{``Fly-swat or cannon''} (\framework{}). Given a set of inputs and a set of candidate LMs, \framework{} judiciously assigns each input to an LM predicted to do well on the input according to a so-called meta-model, aiming to achieve high overall performance at low cost.
The cost--performance tradeoff can be flexibly tuned by the user.
Options include, among others, maximizing total expected performance (or the number of processed inputs) while staying within a given cost budget, or minimizing total cost while processing all inputs.
We evaluate \framework{} on 14 datasets covering five natural language tasks, using four candidate LMs of vastly different size and cost.
With \framework, we match the performance of the largest available LM while achieving a cost reduction of 63\%.
Via our publicly available library,%
\footnote{https://github.com/epfl-dlab/forc}
researchers as well as practitioners can thus save large amounts of money without sacrificing performance.

\end{abstract}




\maketitle

\section{Introduction}
\begin{figure}
    \centering
    \includegraphics[width=\columnwidth]{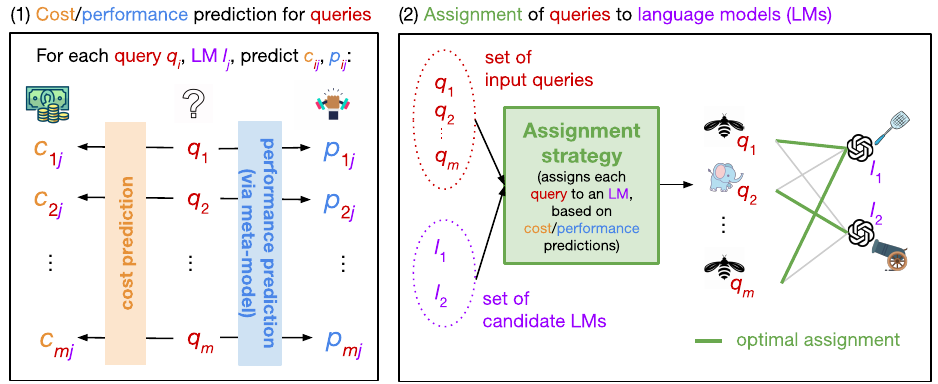}
    \vspace{-3mm}
    \caption{Overview of \framework,
    our framework for cost-effective LM choice (details in \Secref{sec:framework}).
    \framework{} consists of two steps:
    (1)~Predict cost and performance of each candidate LM on each input query. Cost prediction is done using API pricing. Performance prediction is done using a \textit{meta-model,} trained ahead of time (not shown) based on existing pairs of LM queries and LM performance scores.
    (2)~Assign each query to at most one LM using an assignment strategy, aiming for high total expected performance at low cost.
    Note that neither of the two steps requires interacting with the LMs; queries are fed to the assigned LMs only after the above steps.}
    \label{fig:framework_diagram}
    \vspace{-2mm}
\end{figure}
In recent years, a clear trend has emerged in natural language processing and has subsequently spread across data science, characterized by the increasing prominence of large language models (LLMs). With the wide range of applications these models are capable of solving, many companies have contributed to this trend by offering their own LLMs as a service. As a result, the landscape of language processing is undergoing a dynamic shift, with an increasing number of LLMs becoming available on the market.

As the size of LLMs continues to expand, their capabilities are undergoing substantial enhancements, leading to notable improvements across various language-related tasks ~\cite{chowdhery2022palm, brown2020language, touvron2023llama, hoffmann2022training}. With the growth in the number of parameters, their ability to understand complex contexts has improved significantly. These bigger models are better at picking up subtle changes in meaning, which helps them give more relevant responses that fit the context. Moreover, their increased size allows LLMs to generate text that is more coherent and fluent, often resembling a human-like conversation. Finally, as training datasets have been getting larger, the amount of knowledge baked into LLM parameters has also broadened. This results in improved factual accuracy and a better capability to provide well-informed answers to a wider range of questions.

However, as the use of LLMs becomes more common, there is a relevant concern about the rising costs of running them. State-of-the-art language models (LMs) have hundreds of billions of parameters, so they need much computing power, leading to higher expenses. For instance, running GPT-4 with an 8K-token context is 20 times more expensive than running GPT-3.5 on the same query with a 4K-token context.\footnote{\url{https://openai.com/pricing}}
Even though LLMs excel at handling complex language tasks, it is important to realize that not every situation needs their massive capabilities. Smaller LMs are generally good at handling simpler language tasks and can be a more cost-effective choice in cases where full LLM power is not necessary. For example, on the 14 datasets that we examined with four different language models, 33\% of data samples are successfully solved both by the biggest model and at least one of the smaller ones, while 11\% are exclusively solved by one or more of the smaller models, with the biggest model failing to answer correctly (\cf\ \Secref{sec:framework_results}).

There is, thus, an opportunity to save cost by assigning each input to the cheapest model able to solve it. The problem in realizing this is how to predict ahead of time which models would correctly solve which inputs---without actually running each LM on each input, which would defeat the purpose.
\citet{chen2023frugalgpt} proposed to employ increasingly expensive LMs in a cascade until a satisfactory result is obtained.
This requires querying potentially multiple LMs per input, something we set out to avoid in our approach.

\xhdr{Proposed solution}
In this paper, we propose a novel approach for saving LM costs by introducing \textit{``Fly-swat or cannon''} (\framework{}), a cost-aware framework that aims to assign each query from a query set provided by the user to an appropriate LM, without the need to run any of the LMs in the process. As shown in~\Figref{fig:framework_diagram},
\framework{} consists of two steps:
First, we predict the cost and performance of each candidate LM on each input query. Cost prediction is done using the LM provider's API pricing; performance prediction is done using a \textit{meta-model,} a regression model trained ahead of time based on existing pairs
of LM queries and LM performance scores.
Second, we assign each query to at most one LM using an assignment strategy, aiming for high total expected performance at low cost. The cost--performance tradeoff can be tuned by choosing from multiple strategies, each formalized as an optimization problem.



\xhdr{Advantages of \framework}
As mentioned, \framework\ does not require any interaction with the LMs when assigning queries to LMs.
Rather, each query is fed to its assigned LM only once \framework{} has terminated.
This opens up the possibility for greater budget savings in comparison to existing work.
Next, the meta-model can be trained on inputs from the union of a wide range of tasks and datasets, without using any information about the task or dataset from which an input was sourced.
This way, at run time, \framework\ can handle inputs without having to know which tasks they correspond to, and as we show, \framework{} even works on inputs from tasks not seen during meta-model training.
Finally, compared to prior work, we offer users more flexibility by providing them with more options for cost and performance constraints and preferences.

\xhdr{Results}
With the help of \framework, on 14 datasets spanning five different task types, with four different LMs available, we are able to reduce the cost of running the test dataset by 63\% while maintaining the same performance as the biggest LM. To facilitate the use of \framework, we release the library as open-source code.\footnotemark[1]

\xhdr{Contributions}
Briefly, our contributions are the following:




\begin{enumerate}
    \item We propose \framework, a cost-aware framework that automatically assigns input queries to suitable LMs without the need to run the LMs themselves in the process.
    \item We show that, by employing \framework, we are able to substantially reduce the cost of running queries from different tasks, while maintaining performance equal to the biggest LM available in our evaluation.
    \item We release a library for our framework, enabling users to run the setting introduced here or to train different meta-models tailored to their needs.
\end{enumerate}

    
    


\section{Background and Related Work}
\subsection{Language-model evaluation}

Thorough assessment of LMs is a complex, but necessary task required for investigating and improving LM performance. With the appearance of general purpose LMs, the need for an all\hyp encompassing evaluation across different tasks to set a common standard became apparent. There have been several efforts to simplify this evaluation process. For instance, EleutherAI's Language Model Harness Evaluation framework~\cite{eval-harness}, Huggingface's Evaluate library~\cite{vonwerra2022evaluate}, and BIG-Bench~\cite{srivastava2023imitation} all offer convenient open-source repositories that enable common evaluation and encourage collaborative advancements in the field.

\citet{liang2022holistic} performed an exhaustive evaluation of a wide set of LMs, termed \textit{Holistic Evaluation of Language Models} (HELM). They evaluated LMs of different sizes and capabilities, on various datasets and tasks, from many performance aspects, such as accuracy, robustness, fairness, etc. This enabled a standardized view of LM performance and a more reliable way to compare LMs.

Their results revealed that smaller LMs are able to solve some tasks as well as bigger LMs. This is an indicator that it is unnecessary to use the biggest LM for each scenario. However, for more complex tasks, as smaller LMs mostly fail to solve them, we should use bigger, more capable LMs. These insights demonstrate that there is a need for an automatized framework that would help us decide when to use which LM.
The results align with the findings from the Inverse Scaling Prize competition~\cite{mckenzie2022inverse}. Results from the two rounds of competition~\cite{mckenzie2022round1, mckenzie2022round2} revealed numerous tasks where larger LMs perform worse than their smaller counterparts.





\subsection{Inference-cost optimization}
The majority of the cost for an LM product comes from inference, not training~\cite{touvron2023llama}. Despite this, most existing research focuses on minimizing training cost~\cite{hoffmann2022training, kaplan2020scaling, touvron2023llama}.
The high cost of executing these models has driven the development of several inference cost reduction techniques, such as quantization~\cite{gholami2021survey, NEURIPS2022_096347b4, pmlr-v202-xiao23c}, distillation~\cite{Gou_2021,sanh2020distilbert, jiao-etal-2020-tinybert}, and pruning~\cite{kurtic2023ziplm,JMLR:v22:21-0366}.

\citet{zong2023model} include not only the training cost, but also annotation and inference cost in their empirical analysis. They focus on one type of task only---text classification---and evaluate different types of models, including non-neural models, an LLM, and smaller language models. They give insights on which model would be the best choice to train or use in a specific real-world scenario. Contrary to this, our work focuses on LMs only, developing a framework that automatically decides which LM is suitable for the tasks presented under inference budget constraints.

Similarly to our work, \citet{chen2023frugalgpt} attempt to develop a framework working with LMs only. They propose using LMs in cascade, i.e., sending a query to available LMs sequentially until the reliability of an answer is over some predefined threshold. Their approach is specialized, meaning that parts of the framework are adapted specifically for one dataset during training. We, on the contrary, focus on determining which LM to use prior to sending queries to any of them. As we do not need to run any of the LMs to find the best one, our approach is potentially a lot cheaper. In addition, we tested it on a wider range of datasets, and we developped a general framework without the need to retrain our meta-model for each dataset separately.


    
    
    
    
    



\section{Method}

\begin{table}
\centering
\resizebox{\columnwidth}{!}{
\setlength{\tabcolsep}{5pt}
\begin{tabular}{@{}llllll@{}}
\toprule
Datset          & Task type           & Evaluation metric       & Train size & Val size & Test size \\ \midrule \midrule
MMLU            & QA                  & Exact match (EM)        & 434        & 47       & 86        \\
RAFT            & Text classification & Quasi-exact match       & 341        & 33       & 66        \\
WikiFact        & QA                  & Quasi-exact match       & 6069       & 669      & 1189      \\
BoolQ           & QA                  & Quasi-exact match       & 765        & 85       & 150       \\
TruthfulQA      & QA                  & Exact match (EM)        & 500        & 56       & 98        \\
IMDB            & Sentiment analysis  & Quasi-exact match       & 765        & 85       & 150       \\
Entity matching & Reasoning           & Quasi-exact match       & 1071       & 119      & 210       \\
Data imputation & Reasoning           & Quasi-exact match       & 324        & 37       & 63        \\
bAbI            & Reasoning           & Quasi-exact match       & 765        & 85       & 150       \\
MATH            & Reasoning           & Equivalent              & 334        & 37       & 66        \\
GSM8K           & Reasoning           & EM (up to an indicator) & 765        & 85       & 150       \\
LSAT            & Reasoning           & Quasi-exact match       & 353        & 39       & 69        \\
LegalSupport    & Reasoning           & Quasi-exact match       & 765        & 85       & 150       \\
CivilComments   & Toxicity detection  & Quasi-exact match       & 765        & 85       & 150       \\ 
\midrule
Total & - & - & 14016 & 1547 & 2747 \\

\bottomrule
\end{tabular}
}
\caption{Dataset specification.}
\label{tab:datasets_specs}
\vspace{-5mm}
\end{table}
\subsection{Problem setting}
\xhdr{Tasks}
We assume that the user has a set of queries they want to solve using an LM. A lot of the commonly known tasks, such as question-answering (QA), reasoning, and summarization, can be written as textual queries, or prompts. For example, to transform a summarization task into query format, one can construct an LM prompt by adding an instruction such as ``Summarize the above article in one sentence'' to the text to be summarized. Even simpler, QA tasks can often be sent to the LM in their original form.

We work under the premise that all of the queries are evaluated using the same, single metric. This way, we can be sure that the evaluation process is straightforward and consistent across all tasks.


    

\xhdr{LMs}
We consider a scenario where a user has access to $k$ LMs ($l_i$ for $i=1,\ \dotsc,\ k$) that could potentially solve the set of $m$ queries ($q_j$ for $j=1,\ \dotsc,\ m$). These LMs can be different in their capabilities and size. Each LM is associated with its own cost. Costs can be defined by the user, e.g., via the LMs' API pricing.

\xhdr{Goal}
Our goal is to use the pool of LMs to solve queries in a budget-conscious way. We aim to assign each query $q_j$ to at most one $l_i$ while respecting the user's cost--performance requirements.

\subsection{Framework setting}
\label{sec:framework}
Our framework has three main components: \modelname, cost estimation, and assignment strategy. In~\Figref{fig:framework_diagram}, we illustrate the way \framework\ works. First, the user needs to specify a set of queries they want to solve. Then, using the \modelname, we predict the performance $p_{ij}$ of each LM $l_i$ on each query $q_j$. At the same time, we estimate the cost $c_{ij}$ of query $q_{j}$ when using LM $l_{i}$. Next, the user needs to specify one of the assignment strategies described below, and optional cost--performance requirements. The strategy will then be used to assign each query to at most one of the LMs.

\xhdr{Meta-model and cost estimation}
In order to know which LM to use for a certain query, we first have to predict the performance $p_{ij}$ that LM $l_i$ would achieve on query $q_j$. To do so, we train a \modelname. During training, we send a query $q_j$, plus a token representing $l_i$, as input to the \modelname, with target output $p_{ij}$, which we assume is known for the training set. Our \modelname\ is significantly smaller than all the LMs we are working with (\cf\ \Secref{sec:exp_lms}) and it is trained on a diverse set of tasks, which allows it to be suitable for general use (\cf\ \Secref{sec:data}).
It is worth noting that one can train a \modelname\ tailored to one's own needs, with different LMs in the pool, different datasets, or different performance metrics, and plug it into the framework pipeline.

Along with the measure of performance, we have to estimate the cost $c_{ij}$ of query $q_j$ on each LM $l_i$. 
The cost function can be customized as needed.
For implementation details, see~\Secref{sec:exp_lms}.

\xhdr{Assignment strategies}
\label{sec:strategies}
Once we have a \modelname, we need to decide how to assign each query to one of the possible LMs in the pool, based on performance and cost estimates. We call the method to do this an \emph{assignment strategy}. There are two types of strategies:

\noindent (i) \textbf{Cost-insensitive strategies}: When applying cost-insensitive strategies to the samples, we do not consider any constraints on the budget or performance that the user might have set. Each data sample is treated in the same way, independently of the whole batch. We define the following cost-insensitive strategies:

        (a) \textit{Single-model strategy}: This strategy implies applying a single, fixed LM from the available LMs to each sample.

        (b) \textit{Performance-maximizing strategy}: This strategy is based on the outputs of the \modelname. For each sample, we choose the LM that, according to the \modelname, is predicted to achieve the highest performance.

        (c) \textit{Thresholding strategy}: This strategy is also based on the outputs of the \modelname. The user has to specify an acceptable\hyp performance threshold that defines whether a task is solved or not. Outputs are binarized according to that threshold. A concrete example where this strategy might be useful are tasks that are evaluated with binary metrics such as accuracy. The strategy works by choosing the cheapest LM that solves the respective data sample. In cases where none of the LMs solve the sample according to our \modelname, we examine two possibilities: choosing the smallest (and thus generally cheapest) LM, or choosing the biggest (and thus generally most powerful) LM for that data sample.


    \noindent (ii) \textbf{Cost-sensitive strategies}: Contrary to cost-insensitive strategies, in this setting, we consider constraints, such as cost constraints, set by the user for the batch of data samples in its entirety. This transforms the problem into an optimization problem. We employ the following cost-sensitive strategies:
    
        (a) \textit{Cost-oriented ILP strategy}: We formulate the problem of assigning an LM to each sample as an integer linear programming (ILP) problem. We define $M$ as a set of LMs, $S$ as a set of samples that need to be assigned to LMs, and $C_{\max}$ as the maximum total cost of processing all the samples. A binary variable $x_{ij}$ is introduced to describe the assignment (or lack of it) between a data sample $q_j$ and an LM $l_i$. If $x_{ij} = 1$, sample $q_j$ is assigned to the LM $l_i$. A sample does not necessarily have to be assigned to any LM. Assigning sample $q_j$ to LM $l_i$ is associated with cost $c_{ij}$ and value $p_{ij}$, where cost $c_{ij}$ corresponds to the estimated cost, and value $p_{ij}$ corresponds to the predicted performance when using LM $l_i$ to solve the sample $q_j$. The goal is to maximize the performance on the whole set of samples while respecting the cost constraint. This problem is then formalized as an ILP as follows:
        \begin{align}
            \text{maximize} && \sum_{l_i\in M, q_j\in S} p_{ij}\,x_{ij} \label{eq:objective} \\
            \text{s.t.} && \sum_{l_i\in M} x_{ij} \leq 1, \;\;\; \forall q_j \in S \label{eq:matching_constraint} \\
            && \sum_{l_i\in M, q_j\in S} c_{ij}\, x_{ij} \leq C_{\max} \label{eq:cost_constraint}
        \end{align}
        Here, (\ref{eq:matching_constraint}) ensures that every sample is assigned to at most one LM.

        (b) \textit{Performance-oriented ILP strategy}: Similar to the previous case, we formulate this problem in the form of an ILP. The goal in this case is to minimize the cost, while respecting the performance constraint $P_{\min}$ that the user has set. In short, following the same notation as before, this problem is formalized as follows:
        \begin{align}
            \text{minimize} && \sum_{l_i\in M, q_j\in S} c_{ij}\,x_{ij} \label{eq:objective_perf} \\
            \text{s.t.} && \sum_{l_i\in M} x_{ij} \leq 1, \;\;\; \forall q_j \in S \label{eq:matching_constraint_perf} \\
            && \sum_{l_i\in M, q_j\in S} p_{ij} \,x_{ij} \geq P_{\min} \label{eq:perf_constraint}
        \end{align}
        This strategy can also be implemented when binarizing the performance values $p_{ij}$, as done under the cost-insensitive thresholding strategy. In that case, the performance-oriented ILP strategy can be viewed as minimizing the cost for solving at least $P_{\min}$ samples.

        (c) \textit{Greedy strategy}: This strategy works by going through the samples sequentially and choosing, for each sample, the LM achieving the highest performance according to the \modelname, until the cost constraint is reached. After this point, remaining data samples remain unassigned and are not fed to any of the LMs in the pool. For our experiments (\cf\ \Secref{sec:framework_eval}), they are counted as incorrect (when using accuracy as the performance metric), with cost zero.

\section{Experiments}
\subsection{Meta-model evaluation}

\begin{table}
\centering
\resizebox{\columnwidth}{!}{
\setlength{\tabcolsep}{5pt}
\begin{tabular}{@{}lllll@{}}
\toprule
                        & text-ada-001 & text-babbage-001 & text-curie-001 & text-davinci-002 \\ \midrule
Pricing (\$ per 1k tokens) & 0.0004       & 0.0005           & 0.002          & 0.02             \\
Average output length   & 6.85         & 7.18             & 7.01           & 8.41             \\ \bottomrule
\end{tabular}
}
\caption{Specifications of available LMs.}
\label{tab:lm_specs}
\vspace{-5mm}
\end{table}

\xhdr{Data}
\label{sec:data}
As our main source of data, we use raw LM runs from the HELM project \cite{liang2022holistic}. Raw runs consist of inputs (queries and full prompts) sent to the LM, generation parameters, ground truth references, and LM outputs with additional details such as log probability and the time it took to execute the prompt.

Raw model runs have been released for a wide range of datasets covering multiple tasks. While all of these tasks are in the same input format (query), standard metrics used to evaluate the quality of the output can be vastly different between tasks. For example, summarization output is often evaluated using ROUGE scores~\cite{lin-2004-rouge}, which are continuous values from 0 to 1, and question-answering can be evaluated using EM (exact match), which is a binary metric. 

For the sake of this paper, we focus on tasks for which there is a clear answer to whether the model's output solves the query or not. This means we focus on tasks that are normally evaluated only using binary metrics. In particular, the datasets we are working with are evaluated with one of the following metrics:




\begin{itemize}
    \item \textbf{Exact match (EM)}: The LM output matches the ground-truth reference string exactly.
    \item \textbf{Quasi-exact match}: As defined by \citet{liang2022holistic}, this metric equals EM, but after slightly preprocessing output and ground truth (e.g., by lower-casing, removing whitespace, punctuation, and articles).
    \item \textbf{Equivalent}: LM output has to be mathematically equal to the ground truth reference.
\end{itemize}
Following the terminology introduced by~\citet{liang2022holistic}, we refer to all of these metrics, applied to the whole set of samples, as \emph{accuracy}. For more details on datasets and types of tasks, see~\Tabref{tab:datasets_specs}.

To train and evaluate the \modelname, we use LM queries as inputs, and as outputs we use the performance scores calculated based on the ground-truth references and the LM's outputs. The metric used to calculate the score is dependent on the task and dataset that the query comes from (see~\Tabref{tab:datasets_specs}). We append an LM token ``[LM$_i$]'' to the input to differentiate for which LM we are estimating the probability of answering the query.

\xhdr{LMs}
\label{sec:exp_lms}
We opt to work with a set of OpenAI models tested by ~\citet{liang2022holistic}, for two reasons. First, while the size of these LMs is not publicly available, all of them differ in their capabilities.\footnote{\url{https://platform.openai.com/docs/models/overview}} This diversity provides our framework with the flexibility to employ less powerful models for simpler queries and more potent models for more complex ones.
Second, for this set of models, there is a fairly straightforward way to calculate the cost of each sample as the price of running the query with the selected LM using the OpenAI API. For the cost of the output, which is unknown prior to running the respective query, we calculate the average length of the outputs (in tokens) as available in the HELM data and apply the same pricing as for the query. For details on pricing at the time of training and average output lengths, see~\Tabref{tab:lm_specs}.

\xhdr{Implementation}
\label{sec:implementation}
Our \modelname\ is a DistilBERT model\footnote{Initialized with the weights of the 'distilbert-base-uncased' model; see \url{https://huggingface.co/distilbert-base-uncased}.} (66M parameters), finetuned on the collected dataset of raw runs. It was trained using the Adam optimizer with learning rate $3 \times 10^{-5}$ and 0.1 gradient clipping on the Euclidean norm. The model was trained for 3,000 steps with batch size 16 and a polynomial learning rate scheduler with a final learning rate of 0. Training was performed on a machine with a single Tesla T4 16GB GPU, taking around 2h.

As a baseline to compare against, we use a dummy classifier that always predicts the most frequent class, depending on the dataset that the query comes from. It is worth noting that, during inference, the \modelname\ works only with the query, without the need to specify the dataset from which the query comes.

\xhdr{Evaluation metrics}
To evaluate the performance of the \modelname, we use standard metrics: accuracy, precision, recall, and F1 score. We calculate macro scores to give equal weight to each class, regardless of its frequency. We additionally calculate ROC-AUC and PR-AUC scores. To distinguish between the accuracy of the \modelname\ and that of the actual language models themselves, we refer to the \modelname{}'s accuracy as \emph{meta-accuracy}. 

\begin{table*}[t]
\centering
\resizebox{\textwidth}{!}{
\setlength{\tabcolsep}{5pt}
\begin{tabular}{lllllll|llll}
\toprule
& \multicolumn{6}{c}{Meta-model} & \multicolumn{4}{c}{Dummy classifier (majority label per dataset)}\\
 & Meta-accuracy & Precision & Recall & F1 & ROC-AUC & PR-AUC & Meta-accuracy & Precision & Recall & F1\\
\midrule
\midrule
\emph{\textbf{Overall}} & 81.57* {\scriptsize± 0.78} & 79.90* {\scriptsize± 0.81} & 80.65* {\scriptsize± 0.64} & 80.26* {\scriptsize± 0.75} & 80.62* {\scriptsize± 0.83} & 79.30* {\scriptsize± 0.85} & 79.60* {\scriptsize± 0.69} & 78.03* {\scriptsize± 0.89} & 77.25* {\scriptsize± 0.79} & 77.62* {\scriptsize± 0.71} \\
\midrule
\multicolumn{3}{l}{\emph{\textbf{Dataset}}} \\
\hspace{4mm} MMLU & 73.08* {\scriptsize± 3.63} & 69.22* {\scriptsize± 6.48} & 61.63* {\scriptsize± 4.64} & 61.89* {\scriptsize± 5.25} & 61.96* {\scriptsize± 4.37} & 58.47 {\scriptsize± 9.09} & 68.59* {\scriptsize± 4.81} & 34.32* {\scriptsize± 2.51} & 50.00* {\scriptsize± 0.00} & 40.69* {\scriptsize± 1.51} \\
\hspace{4mm} RAFT & 67.59* {\scriptsize± 5.37} & 68.98* {\scriptsize± 5.88} & 65.64* {\scriptsize± 5.52} & 65.76* {\scriptsize± 5.42} & 65.83* {\scriptsize± 5.37} & 79.60 {\scriptsize± 4.12} & 55.33* {\scriptsize± 5.92} & 27.71* {\scriptsize± 2.61} & 50.00* {\scriptsize± 0.00} & 35.48* {\scriptsize± 2.52} \\
\hspace{4mm} WikiFact & 87.21 {\scriptsize± 0.82} & 71.06* {\scriptsize± 2.50} & 62.25* {\scriptsize± 1.77} & 64.66* {\scriptsize± 1.98} & 62.36* {\scriptsize± 1.85} & 44.94* {\scriptsize± 3.99} & 86.95 {\scriptsize± 0.95} & 43.47* {\scriptsize± 0.47} & 50.00* {\scriptsize± 0.00} & 46.52* {\scriptsize± 0.25} \\
\hspace{4mm} Entity matching & 90.00* {\scriptsize± 1.86} & 94.87* {\scriptsize± 1.04} & 57.67* {\scriptsize± 3.44} & 60.36* {\scriptsize± 4.73} & 57.25* {\scriptsize± 3.35} & 94.83* {\scriptsize± 1.15} & 88.10* {\scriptsize± 2.07} & 44.05* {\scriptsize± 0.93} & 50.00* {\scriptsize± 0.00} & 46.83* {\scriptsize± 0.65} \\
\hspace{4mm} Data imputation & 81.02* {\scriptsize± 4.78} & 77.10* {\scriptsize± 7.17} & 70.32* {\scriptsize± 5.79} & 71.90* {\scriptsize± 7.00} & 69.31* {\scriptsize± 6.17} & 90.75* {\scriptsize± 2.94} & 73.92* {\scriptsize± 5.47} & 37.07* {\scriptsize± 2.44} & 50.00* {\scriptsize± 0.00} & 42.41* {\scriptsize± 1.74} \\
\hspace{4mm} BoolQ & 64.82* {\scriptsize± 3.55} & 65.33* {\scriptsize± 5.19} & 56.93* {\scriptsize± 2.83} & 54.07* {\scriptsize± 4.15} & 56.94* {\scriptsize± 2.66} & 80.92 {\scriptsize± 2.05} & 60.85* {\scriptsize± 3.49} & 30.47* {\scriptsize± 2.05} & 50.00* {\scriptsize± 0.00} & 37.84* {\scriptsize± 1.52} \\
\hspace{4mm} TruthfulQA & 76.02* {\scriptsize± 4.06} & 70.48* {\scriptsize± 5.08} & 68.05* {\scriptsize± 5.12} & 68.88* {\scriptsize± 4.79} & 67.81* {\scriptsize± 4.98} & 61.79 {\scriptsize± 7.87} & 71.51 {\scriptsize± 4.50} & 35.57* {\scriptsize± 2.16} & 50.00* {\scriptsize± 0.00} & 41.55* {\scriptsize± 1.67} \\
\hspace{4mm} IMDB & 86.35* {\scriptsize± 2.71} & 43.42* {\scriptsize± 1.42} & 49.49* {\scriptsize± 0.39} & 46.30* {\scriptsize± 0.74} & 49.51* {\scriptsize± 0.43} & 93.48* {\scriptsize± 1.07} & 86.90* {\scriptsize± 2.67} & 43.58* {\scriptsize± 1.36} & 50.00* {\scriptsize± 0.00} & 46.54* {\scriptsize± 0.87} \\
\hspace{4mm} bAbI & 74.57* {\scriptsize± 3.72} & 73.48* {\scriptsize± 3.22} & 74.11* {\scriptsize± 3.55} & 73.78* {\scriptsize± 3.83} & 73.95* {\scriptsize± 3.51} & 75.18* {\scriptsize± 3.64} & 59.16* {\scriptsize± 4.26} & 29.63* {\scriptsize± 1.91} & 50.00* {\scriptsize± 0.00} & 37.18* {\scriptsize± 1.45} \\
\hspace{4mm} MATH & 92.89 {\scriptsize± 3.19} & 75.07 {\scriptsize± 26.13} & 52.42 {\scriptsize± 3.57} & 52.91 {\scriptsize± 8.01} & 52.38 {\scriptsize± 4.01} & 56.43 {\scriptsize± 5.11} & 92.56 {\scriptsize± 2.84} & 46.05 {\scriptsize± 1.45} & 50.00 {\scriptsize± 0.00} & 48.01 {\scriptsize± 0.84} \\
\hspace{4mm} GSM8K & 90.82 {\scriptsize± 1.77} & 45.51 {\scriptsize± 0.89} & 50.00 {\scriptsize± 0.00} & 47.67 {\scriptsize± 0.61} & 50.00 {\scriptsize± 0.00} & 54.50 {\scriptsize± 1.11} & 91.05 {\scriptsize± 1.99} & 45.54 {\scriptsize± 0.90} & 50.00 {\scriptsize± 0.00} & 47.64 {\scriptsize± 0.58} \\
\hspace{4mm} LSAT & 76.83 {\scriptsize± 4.58} & 38.48 {\scriptsize± 2.13} & 50.00 {\scriptsize± 0.00} & 43.37 {\scriptsize± 1.63} & 50.00 {\scriptsize± 0.00} & 61.54 {\scriptsize± 2.38} & 77.19 {\scriptsize± 4.68} & 38.28 {\scriptsize± 2.56} & 50.00 {\scriptsize± 0.00} & 43.60 {\scriptsize± 1.68} \\
\hspace{4mm} LegalSupport & 49.45 {\scriptsize± 3.30} & 48.97* {\scriptsize± 5.16} & 49.30 {\scriptsize± 2.99} & 44.17* {\scriptsize± 3.30} & 49.46 {\scriptsize± 3.03} & 70.06* {\scriptsize± 2.78} & 50.58 {\scriptsize± 4.55} & 25.28* {\scriptsize± 1.91} & 50.00 {\scriptsize± 0.00} & 33.61* {\scriptsize± 1.57} \\
\hspace{4mm} CivilComments & 79.22 {\scriptsize± 3.01} & 39.57 {\scriptsize± 1.80} & 50.00 {\scriptsize± 0.00} & 44.17 {\scriptsize± 0.99} & 50.00 {\scriptsize± 0.00} & 89.58 {\scriptsize± 1.64} & 79.04 {\scriptsize± 2.88} & 39.68 {\scriptsize± 1.48} & 50.00 {\scriptsize± 0.00} & 44.10 {\scriptsize± 0.98} \\
\midrule
\multicolumn{3}{l}{\emph{\textbf{Task}}} \\
\hspace{4mm} QA & 83.40* {\scriptsize± 0.88} & 74.09 {\scriptsize± 1.48} & 71.00* {\scriptsize± 1.56} & 72.25* {\scriptsize± 1.40} & 70.96* {\scriptsize± 1.33} & 60.07* {\scriptsize± 2.30} & 82.32* {\scriptsize± 0.74} & 72.85 {\scriptsize± 1.91} & 62.73* {\scriptsize± 1.33} & 65.06* {\scriptsize± 1.59} \\
\hspace{4mm} Sentiment analysis & 86.34* {\scriptsize± 2.50} & 43.48* {\scriptsize± 1.31} & 49.51* {\scriptsize± 0.47} & 46.32* {\scriptsize± 0.78} & 49.54* {\scriptsize± 0.43} & 93.45* {\scriptsize± 1.23} & 86.85* {\scriptsize± 2.63} & 43.47* {\scriptsize± 1.21} & 50.00* {\scriptsize± 0.00} & 46.54* {\scriptsize± 0.72} \\
\hspace{4mm} Toxicity detection & 79.27 {\scriptsize± 2.96} & 39.58 {\scriptsize± 1.51} & 50.00 {\scriptsize± 0.00} & 44.23 {\scriptsize± 0.89} & 50.00 {\scriptsize± 0.00} & 89.59 {\scriptsize± 1.47} & 79.07 {\scriptsize± 3.46} & 39.53 {\scriptsize± 1.69} & 50.00 {\scriptsize± 0.00} & 44.23 {\scriptsize± 0.85} \\
\hspace{4mm} Text classification & 68.11* {\scriptsize± 5.08} & 69.00* {\scriptsize± 6.00} & 65.84* {\scriptsize± 5.43} & 65.36* {\scriptsize± 5.77} & 65.17* {\scriptsize± 4.94} & 79.65 {\scriptsize± 3.97} & 55.14* {\scriptsize± 5.79} & 27.73* {\scriptsize± 2.96} & 50.00* {\scriptsize± 0.00} & 35.33* {\scriptsize± 2.49} \\
\hspace{4mm} Reasoning & 74.64* {\scriptsize± 1.67} & 69.80* {\scriptsize± 1.88} & 70.79* {\scriptsize± 2.03} & 70.18* {\scriptsize± 1.89} & 70.72* {\scriptsize± 1.72} & 64.49* {\scriptsize± 2.81} & 70.87* {\scriptsize± 1.58} & 64.26* {\scriptsize± 2.06} & 63.28* {\scriptsize± 2.21} & 63.58* {\scriptsize± 1.96} \\
\midrule
\multicolumn{3}{l}{\emph{\textbf{LM}}} \\
\hspace{4mm} ada & 85.18* {\scriptsize± 1.28} & 82.33* {\scriptsize± 1.56} & 83.16 {\scriptsize± 1.53} & 82.77* {\scriptsize± 1.51} & 83.15 {\scriptsize± 1.47} & 79.41 {\scriptsize± 1.93} & 83.20* {\scriptsize± 1.54} & 80.02* {\scriptsize± 1.55} & 82.01 {\scriptsize± 1.53} & 80.84* {\scriptsize± 1.63} \\
\hspace{4mm} babbage & 84.37 {\scriptsize± 1.42} & 81.67 {\scriptsize± 1.35} & 83.08 {\scriptsize± 1.45} & 82.21 {\scriptsize± 1.67} & 83.13 {\scriptsize± 1.67} & 79.52 {\scriptsize± 1.79} & 83.34 {\scriptsize± 1.40} & 80.63 {\scriptsize± 1.61} & 82.02 {\scriptsize± 1.44} & 81.23 {\scriptsize± 1.55} \\
\hspace{4mm} curie & 80.88* {\scriptsize± 1.54} & 78.83* {\scriptsize± 1.36} & 79.99* {\scriptsize± 1.70} & 79.23* {\scriptsize± 1.68} & 80.02* {\scriptsize± 1.63} & 77.49* {\scriptsize± 1.93} & 79.27* {\scriptsize± 1.48} & 77.10 {\scriptsize± 1.44} & 76.75* {\scriptsize± 1.63} & 77.07* {\scriptsize± 1.61} \\
\hspace{4mm} davinci & 75.84* {\scriptsize± 1.39} & 75.83* {\scriptsize± 1.51} & 75.75* {\scriptsize± 1.54} & 75.83* {\scriptsize± 1.49} & 75.73* {\scriptsize± 1.58} & 81.01* {\scriptsize± 1.54} & 72.44* {\scriptsize± 1.44} & 74.30 {\scriptsize± 1.42} & 71.91* {\scriptsize± 1.61} & 71.52* {\scriptsize± 1.81} \\
\bottomrule
\end{tabular}
}
\caption{Performance of the meta-model. Precision, recall, and F1 are macro-averaged over classes. We use threshold 0.5 for these metrics. The meta-model is compared to a dummy classifier that outputs the most probable class in the dataset from which the sample comes. Results are stratified by dataset, task, and LM and presented with 95\% confidence intervals. We do not report ROC-AUC and PR-AUC for the dummy classifier because the dummy classifier is threshold-independent and therefore these two metrics are not informative for it. Asterisks (*) mark results where the meta-model (left) performs statistically significantly ($p<0.05$) differently from the dummy classifier (right).}
\label{tab:meta_model_perf}
\end{table*}

\begin{table}
    \centering
    \resizebox{\columnwidth}{!}{
    \setlength{\tabcolsep}{5pt}
    \begin{tabular}{lllllll}
    \toprule
    Dataset & Meta-acc. & Precision & Recall & F1 & ROC-AUC & PR-AUC \\
    \midrule
    \midrule
     MMLU & 71.78 {\scriptsize± 4.51} & 67.85 {\scriptsize± 7.80} & 58.29 {\scriptsize± 3.91} & 56.95 {\scriptsize± 5.57} & 57.77 {\scriptsize± 4.34} & 55.12 {\scriptsize± 8.77} \\
     RAFT & 54.26 {\scriptsize± 5.73} & 53.49 {\scriptsize± 5.91} & 52.88 {\scriptsize± 5.01} & 52.71 {\scriptsize± 5.80} & 53.01 {\scriptsize± 5.87} & 70.87 {\scriptsize± 5.37} \\
     WikiFact & 72.80 {\scriptsize± 1.29} & 57.49 {\scriptsize± 1.34} & 62.93 {\scriptsize± 1.96} & 57.54 {\scriptsize± 1.47} & 62.78 {\scriptsize± 1.96} & 39.82 {\scriptsize± 2.37} \\
     Entity matching & 77.00 {\scriptsize± 2.53} & 55.07 {\scriptsize± 2.90} & 57.80 {\scriptsize± 4.50} & 55.58 {\scriptsize± 3.91} & 57.03 {\scriptsize± 4.72} & 94.02 {\scriptsize± 1.36} \\
     Data imputation & 71.50 {\scriptsize± 5.41} & 64.23 {\scriptsize± 6.03} & 66.10 {\scriptsize± 6.82} & 64.63 {\scriptsize± 6.68} & 66.57 {\scriptsize± 6.97} & 88.93 {\scriptsize± 3.04} \\
     BoolQ & 59.22 {\scriptsize± 3.82} & 59.30 {\scriptsize± 3.42} & 59.45 {\scriptsize± 4.06} & 59.23 {\scriptsize± 3.88} & 59.80 {\scriptsize± 3.95} & 76.95 {\scriptsize± 3.38} \\
     TruthfulQA & 75.04 {\scriptsize± 4.02} & 68.97 {\scriptsize± 5.80} & 65.10 {\scriptsize± 4.36} & 66.14 {\scriptsize± 4.88} & 65.32 {\scriptsize± 4.97} & 58.89 {\scriptsize± 8.24} \\
     IMDB & 38.99 {\scriptsize± 3.52} & 52.50 {\scriptsize± 2.67} & 54.34 {\scriptsize± 4.74} & 36.90 {\scriptsize± 4.02} & 54.76 {\scriptsize± 4.80} & 90.75 {\scriptsize± 2.74} \\
     bAbI & 64.38 {\scriptsize± 3.75} & 62.75 {\scriptsize± 4.24} & 60.55 {\scriptsize± 3.85} & 60.33 {\scriptsize± 3.98} & 60.19 {\scriptsize± 4.19} & 61.79 {\scriptsize± 5.14} \\
     MATH & 80.90 {\scriptsize± 4.27} & 58.41 {\scriptsize± 5.52} & 70.96 {\scriptsize± 11.44} & 60.55 {\scriptsize± 8.06} & 71.41 {\scriptsize± 9.83} & 43.11 {\scriptsize± 14.01} \\
     GSM8K & 89.55 {\scriptsize± 2.24} & 65.46 {\scriptsize± 6.91} & 60.36 {\scriptsize± 6.10} & 61.95 {\scriptsize± 6.22} & 60.37 {\scriptsize± 5.76} & 33.90 {\scriptsize± 13.61} \\
     LSAT & 74.18 {\scriptsize± 4.71} & 51.70 {\scriptsize± 11.05} & 50.44 {\scriptsize± 3.83} & 47.64 {\scriptsize± 5.10} & 50.37 {\scriptsize± 3.18} & 26.09 {\scriptsize± 14.69} \\
     LegalSupport & 49.88 {\scriptsize± 3.80} & 49.49 {\scriptsize± 3.77} & 49.65 {\scriptsize± 3.63} & 48.89 {\scriptsize± 3.99} & 49.46 {\scriptsize± 3.73} & 65.85 {\scriptsize± 3.73} \\
     CivilComments & 71.36 {\scriptsize± 3.50} & 41.81 {\scriptsize± 3.46} & 45.99 {\scriptsize± 2.03} & 43.44 {\scriptsize± 2.09} & 46.10 {\scriptsize± 1.61} & 87.78 {\scriptsize± 1.73} \\
     \bottomrule
    \end{tabular}
    }
    \caption{Leave-one-out evaluation. Performance of meta-model on a dataset left-out in training. Each row of the table trains a meta-model on all datasets except the one denoted in the row. Testing is then performed on the denoted dataset.}
    \label{tab:generalization_perf}
\end{table}

\begin{figure*}
    \centering
    \begin{subfigure}{0.39\textwidth}
    \centering
    \includegraphics[width=\textwidth]{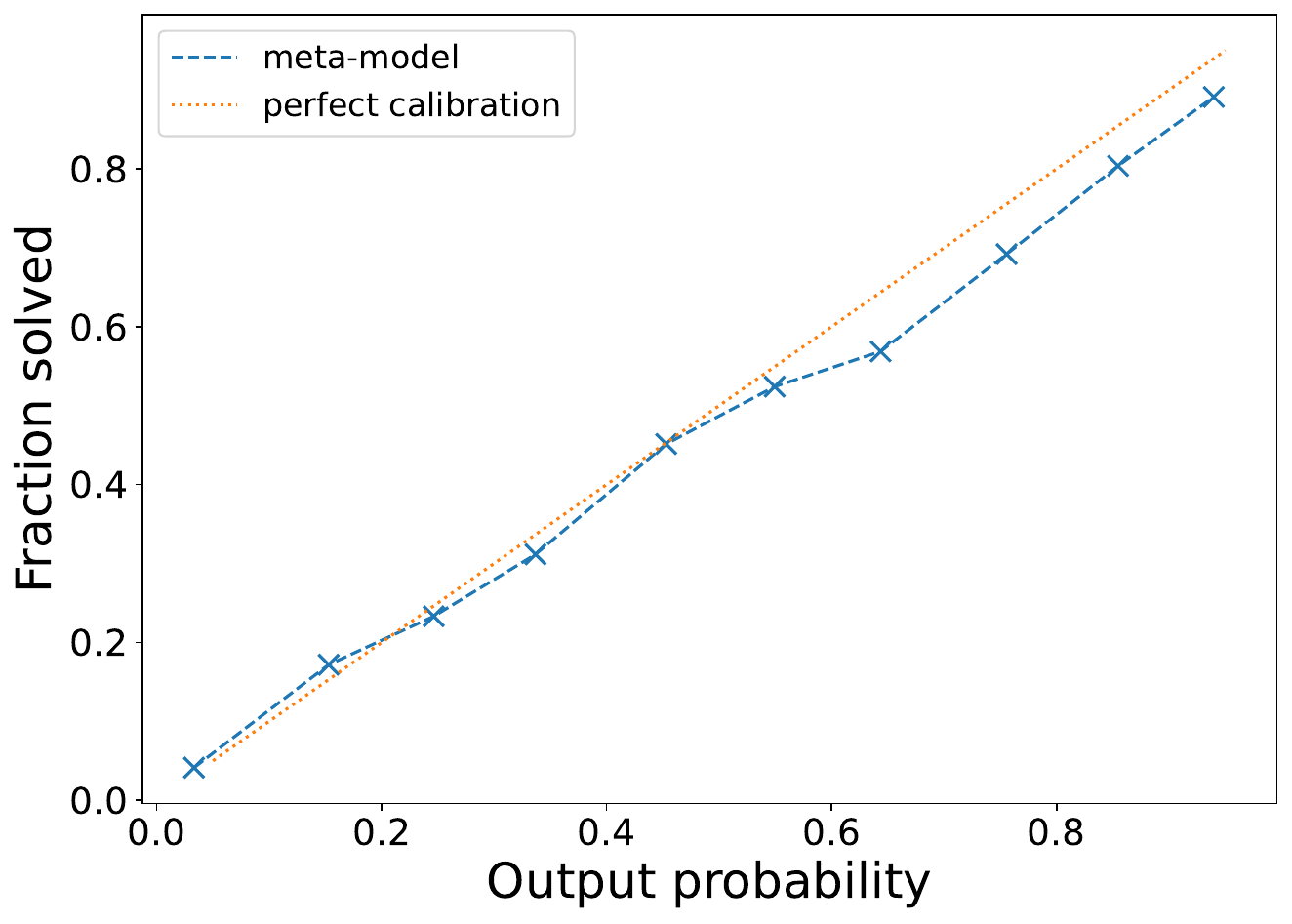}
    \caption{Calibration plot when pooling all 14 datasets}
    \label{fig:calibration_plot_overall}
    \end{subfigure}
    \begin{subfigure}{0.48\textwidth}
    \centering
    \includegraphics[width=\textwidth]{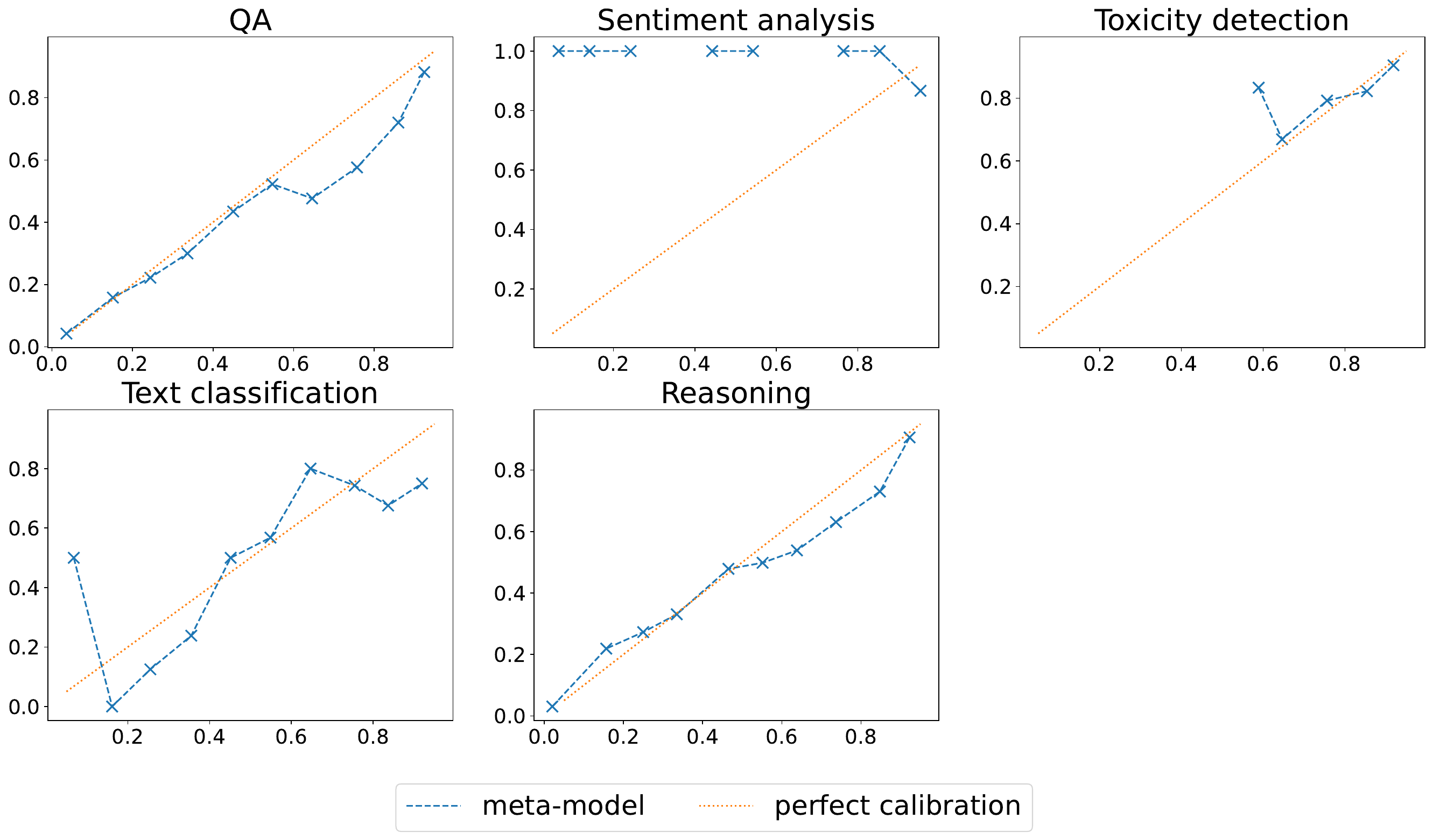}
    \caption{Calibration plots per task}
    \label{fig:calibration_plot_tasks}
    \end{subfigure}
    \vspace{-3mm}
    \caption{Calibration plots. Plots are obtained by grouping the meta-model's output probabilities into bins of equal width and then plotting the fraction of positive samples against the mean value of the bin.}
    \label{fig:calibration_plots}
    \vspace{-2mm}
\end{figure*}

\xhdr{Results}
In~\Tabref{tab:meta_model_perf} we present results of \modelname\ evaluation. Along with results on the whole testing dataset, we present results stratified by datasets, tasks, and LMs.
The overall performance of the \modelname\ is good, as reflected in a high score over all the metrics calculated. The difference between the \modelname\ and the dummy classifier, though significant, is not big. This happens because, as indicated by the meta-accuracy of the dummy classifier (\cf\ column~7 of \Tabref{tab:meta_model_perf}), in most of the datasets the class imbalance is big. As a result, the performance (in terms of meta-accuracy) of the dummy classifier is high. But note that, in a real-world setting, we do not have access to the dataset which the query comes from (it might even correspond to a task not seen during meta-model training), which is essential for the dummy classifier to work. As such, it is not a practical option, but merely an evaluation baseline. Our \modelname{} does not require knowing the task and can be used with arbitrary queries.
Finally, the metric we ultimately care about is not meta-accuracy, but rather the accuracy of the LM chosen based on the meta-model's predictions (as evaluated in \Secref{sec:framework_eval}).


Stratification over datasets and tasks uncovers that there are certain types of datasets and tasks for which it is harder to predict LM performance. These are more complex tasks, which are not trivial for any of the LMs, such as subtypes of the reasoning task.

Stratification over LMs shows that it is harder to predict the behavior of the bigger LMs. Smaller LMs are generally able to work on simpler tasks (for example, see \emph{CivilComments} plot in~\Figref{fig:cost_accuracy_plot_dataset} and \emph{Sentiment analysis} plot in~\Figref{fig:cost_accuracy_plot_task}), but they fail on more complex tasks (see \emph{GSM8K} plot in~\Figref{fig:cost_accuracy_plot_dataset}). Bigger LMs are, on the other hand, able to solve some complex tasks, which means that the \modelname\ has to understand more complicated patterns to correctly predict whether an LM can solve the task or not.

\emph{Calibration.}
We assess calibration via calibration plots obtained by grouping output probabilities into bins of equal width and plotting the fraction of positive samples against the mean value of the bin. In~\Figref{fig:calibration_plot_overall} we can see that the \modelname\ is well calibrated, as the calibration curve lies close to the diagonal representing perfect calibration.
(For task-specific calibration plots, see \Figref{fig:calibration_plot_tasks}.)


\emph{Generalization.}
In an attempt to evaluate how well this \modelname\ would generalize, we retrain it on all datasets but one at a time. We then evaluate the performance of these \modelname{}s on the respective left-out dataset. Results are presented in~\Tabref{tab:generalization_perf}. By comparing these numbers with the ones presented in~\Tabref{tab:meta_model_perf}, we can notice that, although the numbers are generally a little lower, the \modelname\ is able to generalize to the left-out dataset.

\subsection{Framework evaluation}
\label{sec:framework_eval}

\begin{figure}
    \centering
    \includegraphics[width=\columnwidth]{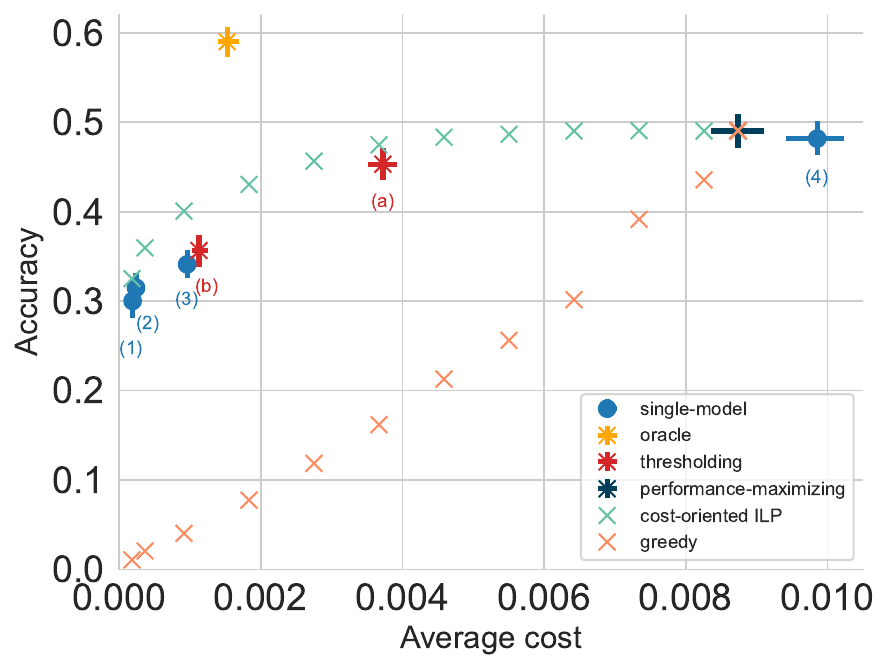}
    \vspace{-4mm}
    \caption{Cost-accuracy plot. Accuracy and average cost per query (in US\$) achieved by assigning every query from the query set to an LM from the LM pool. The plot shows results obtained using assignment strategies from~\Secref{sec:strategies}. Single model strategies for each LMs are marked by the number under them: (1) text-ada-001 (2) text-babbage-001 (3) text-curie-001 (4) text-davinci-002. Two thresholding strategies for cases when none of the LMs solve the data sample are marked by a letter under them: choosing (a) the biggest and (b) the smallest LM. Error bars are 95\% confidence intervals.}
    \label{fig:cost_accuracy_plot}
    \vspace{-3mm}
\end{figure}
\begin{figure*}
    \centering
    \includegraphics[width=0.98\textwidth]{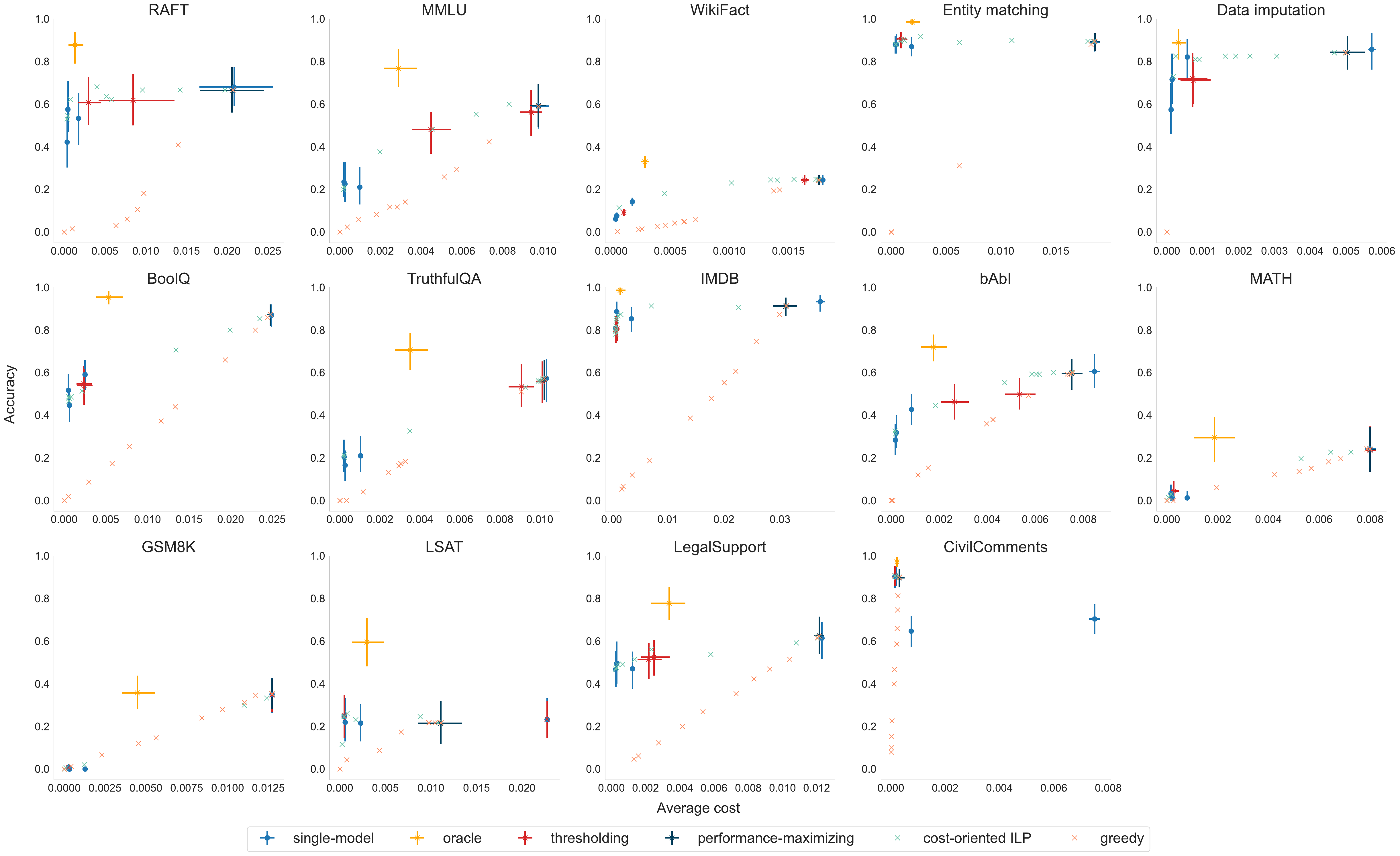}
    \vspace{-3mm}
    \caption{Cost--accuracy plot per dataset. Accuracy and average cost per query (in US\$) achieved by employing strategies from \Secref{sec:strategies} on the test dataset stratified by datasets from \Tabref{tab:datasets_specs}. Single-model and thresholding strategies follow the same trends as in \Figref{fig:cost_accuracy_plot}. Error bars are 95\% confidence intervals.}
    \label{fig:cost_accuracy_plot_dataset}
\end{figure*}
\begin{figure*}
    \centering
    \includegraphics[width=0.98\textwidth]{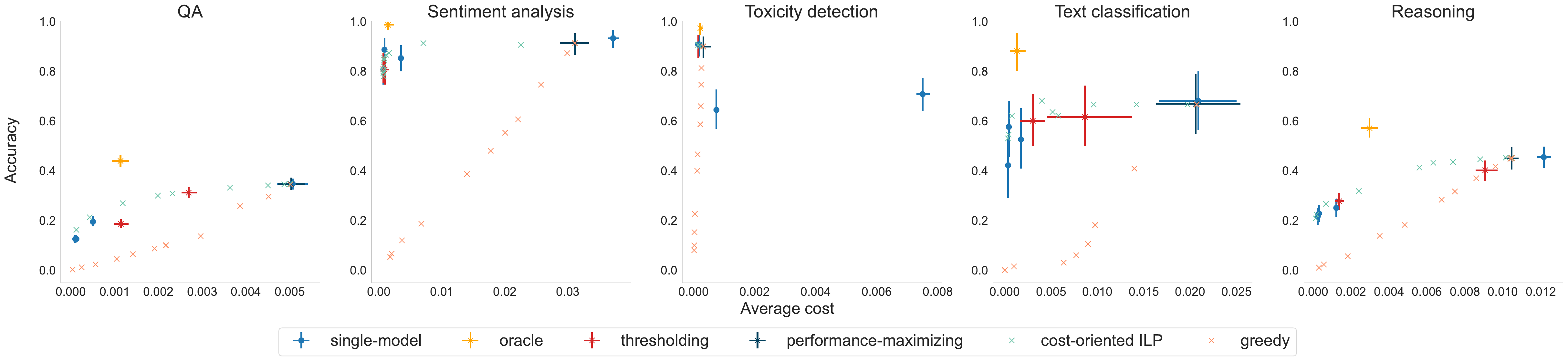}
    \vspace{-2mm}
    \caption{Cost--accuracy plot per task. Accuracy and average cost per query (in US\$) achieved by employing strategies from \Secref{sec:strategies} on the test dataset stratified by tasks from \Tabref{tab:datasets_specs}. Single-model and thresholding strategies follow the same trends as in \Figref{fig:cost_accuracy_plot}. Error bars are 95\% confidence intervals.}
    \label{fig:cost_accuracy_plot_task}
    \vspace{-2mm}
\end{figure*}





\xhdr{Implementation}
In order to evaluate the performance of the framework as a whole, we assign each sample in the test set to one of the LMs in our pool (or none) and calculate the accuracy achieved by running these samples with the assigned LMs. The assignment is done by applying different strategies to the outputs of the \modelname. Additionally, we evaluate cost as described earlier in~\Secref{sec:exp_lms}. Based on those two values, we make a cost--accuracy plot. Cost--accuracy plots (\cf\ \Figref{fig:cost_accuracy_plot}) present the relationship between accuracy achieved using the chosen strategy and the average cost per query needed to perform the run. For cost-sensitive strategies, we run this analysis for a range of cost constraints to trace the cost--performance trade-off.

We opt to evaluate only one ILP based strategy---the cost-oriented one---because the performance-oriented one can be seen as the other side of the same coin. While in a practical sense, the performance-oriented ILP strategy is useful because it offers the user the possibility to estimate the budget needed for the desired quality of the results, this is visible from the cost-oriented strategy results on the cost--accuracy plot when we run it for different cost constraints.

To solve the ILP problem for the cost-oriented strategy, we use the PuLP\footnote{\url{https://coin-or.github.io/pulp/}} library with the \texttt{PULP\_CBC\_CMD} solver. Given the simplicity of the ILP, the total time of assignment for all the queries is overshadowed by the run time of the \modelname. For our testing dataset, obtaining all probabilities from the \modelname\ takes a few minutes, while solving the ILP takes only a fraction of a second. Nonetheless, we leave the option to the user to specify a time limit for ILP execution when working with bigger datasets. If the problem is not solved by the time it reaches the limit, the user will be left with a possibly suboptimal solution.

\xhdr{Oracle}
We calculate the optimal assignment on the ground-truth data by always choosing the cheapest LM that solves the sample. When no LM solves the sample, we assume that the best option is not to send this sample to any of the available LMs. The sample is counted as incorrect when calculating the accuracy, and with a cost of zero when calculating the average cost.


\xhdr{Results}
\label{sec:framework_results}
In~\Figref{fig:cost_accuracy_plot}, we present the results of the framework evaluation in the form of a cost--accuracy plot.
First, we can see that the oracle is not only cheaper than choosing the biggest LM (84.46\% lower cost), but it also performs better (+10.74\% in accuracy). This means that there are cases where a smaller LM performs not only on par, but better than the bigger options. Next, looking at the results under single-model strategies, we can notice a clear difference between the LMs. For comparison, the cheapest LM (text-ada-001) is 98.05\% cheaper than the biggest LM (text-davinci-002), while achieving 18.18\% (absolute) smaller accuracy. There is no significant difference between the two smallest models (text-ada-001 and text-babbage-001), neither in cost nor in accuracy.

Two additional cost-insensitive strategies, the performance\hyp maximizing approach and the thresholding strategy, exhibit comparable performance in terms of accuracy to the largest model (text-davinci-002). Notably, the thresholding strategy stands out for not only its effectiveness but also its much lower cost. In terms of cost effectiveness, it offers a 62.1\% (absolute) reduction relative to text-davinci-002, while the performance-maximizing strategy offers an 11.5\% cost reduction relative to text-davinci-002.

Next, we focus on the cost-sensitive strategies. Note that, for each of these, \Figref{fig:cost_accuracy_plot} shows multiple points, one per maximum available budget (corresponding to the value on the $x$-axis).
By employing cost-sensitive strategies, we are able to further save budget, while achieving essentially the same accuracy as the largest model (text-davinci-002).
In particular, with the cost-oriented ILP strategy, we match text-davinci-002's accuracy for only 37.21\% of the cost. The greedy strategy, on the other hand, performs much worse: in this case, we need 88.57\% of the budget for the same result.

Although the oracle indicates that there is space for improvements in terms of accuracy as well, we did not manage to achieve better accuracy than text-davinci-002 with any budget given to the framework. This happens because our \modelname\ is not able to correctly identify the cases when text-davinci-002 fails to do the task correctly, but some of the smaller models successfully do. This scenario happens in 10.72\% of cases, but we manage to identify it only 0.68\% of the time. For comparison, 33.03\% of data samples are successfully solved by both text-davinci-002 and one of the smaller models, and we correctly recognize this scenario 78.56\% of the time.

In
\Figref{fig:cost_accuracy_plot_dataset}
and
\Figref{fig:cost_accuracy_plot_task}
we present cost--accuracy plots stratified by dataset and task, respectively. First, by focusing on the oracle and text-davinci-002 results on these plots, we spot that, for most of the datasets and tasks, the oracle indicates that there is the same potential to improve both the accuracy and the cost as in the overall case. In rare cases (e.g., \textit{MATH}, \textit{GSM8K}) there is not much space for improvement in terms of accuracy, as small LMs fail to do the task in almost all cases.

Second, patterns similar to the ones present in~\Figref{fig:cost_accuracy_plot} are also visible for most of the datasets and tasks. For example, using the cost-oriented ILP strategy on the question-answering (QA) task, we are able to achieve the same performance as text-davinci-002 for 71.76\% of the price. For the reasoning and the text classification task, these percentages are equal to 83.41\% and 19.15\%, respectively.

For some of the tasks, smaller LMs are also able to solve the queries either in the same way or even better than bigger LMs. As an example of the former, for the sentiment analysis task, text-babbage-001 drops only 4.83\% in accuracy compared to text-davinci-002 for just 2.35\% of its price. In the latter case, on the toxicity detection task, using text-ada-001 results in a 19.87\% jump in accuracy compared to text-davinci-002, while spending only 1.99\% of the budget used for running text-davinci-002.

Thanks to this, depending on the dataset, there are cases where using the \modelname\ helps us achieve both lower cost and higher performance than text-davinci-002. For example, on the \textit{CivilComments} dataset (which is a part of the toxicity detection task), using the cost-oriented ILP strategy results in 20.1\% (absolute) higher accuracy for only 1.99\% of the cost of running text-davinci-002. This result essentially matches the oracle for this dataset.

\section{Discussion}
\subsection{Further use cases of the \framework{} framework} 
\label{sec:framework_use}




In the previous sections (\cf\ \Secref{sec:strategies} and \Secref{sec:framework_eval}), we introduced and evaluated two ILP-based strategies: cost-oriented and performance-oriented. This is, however, not the only potential use of our framework. The user can form an objective function as a custom combination of the terms involving cost and performance, and, similarly, use any combination of the cost and performance constraints. This allows for increased flexibility in practical applications.

Additionally, \framework\ need not be used only to make a decision about an LM to run a certain query. By fixing the LM and calculating the output probabilities for different phrasings of the desired query, one could use the framework to identify the best possible prompt to send to the LM, without the need to spend money by sending them to the LM directly. For this application, it would be advisable to retrain the \modelname\ to fit the task better, as the current \modelname\ was not exposed to small changes in queries during training, and its behavior in such cases is unknown.

\subsection{Practical considerations}
With the emergence of new LMs, our \modelname\ may become stale, as it will not include the possibility to use the newer LMs with likely better abilities. Additionally, as shown by \citet{chen2023chatgpts}, the behavior of GPT-3.5 and GPT-4 is changing over time. While in this paper we do not focus on these two LMs, it is not unreasonable to assume that, with newer versions, the LMs we have been working with exhibit the same behavioral shifts. To minimize the effect of these changes on our framework, \modelname\ should be retrained to take into account any updates in the existing LMs as well as the newly introduced LMs. Because our \modelname\ is small, training is cheap and takes little time, so it is easy to retrain frequently.

In addition, we calculated the costs of running different LMs with the pricing from OpenAI at the time of \modelname\ training. These prices were changing in the past, and will probably change in the future. When such changes happen, the cost parameters should be updated in the framework, as it would affect the cost--accuracy plots and, consequently, the assignment of data samples to LMs.

\section{Conclusion}

In this paper, we address the problem of the increasing costs of running LMs for various tasks. We introduce ``Fly-swat or cannon'' (\framework), a framework for automatically deciding which LM is the best option for a given input query. By employing \framework\ to the union of 14 datasets, we are able to reduce costs by 62.79\% while maintaining the same performance as the biggest LM in our pool. We manage to substantially cut costs on all types of task, and on most of the datasets we evaluate on. We also release a library that enables easy use of our framework, as well as further developments.


\section*{Acknowledgements}
We thank Chris Wendler for helpful feedback.
Robert West's lab is partly supported by grants from
Swiss National Science Foundation (200021\_185043, TMSGI2\_211379),
H2020 (952215),
Microsoft Swiss Joint Research Center,
and Google,
and by generous gifts from Facebook, Google, and Microsoft.

\bibliographystyle{ACM-Reference-Format}
\bibliography{custom}

\end{document}